\documentclass[11pt,letterpaper]{article}

\usepackage[letterpaper, margin=1in]{geometry}
\usepackage[T1]{fontenc}
\usepackage[utf8]{inputenc}

\usepackage{mathptmx}     
\usepackage[scaled=0.92]{helvet} 
\usepackage{courier}      
\usepackage{microtype}

\usepackage{amsmath,amssymb,amsthm,bm}
\usepackage{mathtools}

\usepackage{booktabs}
\usepackage{array}
\usepackage{multirow}
\usepackage{tabularx}
\usepackage{float}
\usepackage{placeins}

\usepackage{graphicx}
\usepackage{caption}
\usepackage{tikz}
\usetikzlibrary{arrows.meta,positioning,fit,backgrounds,calc,shapes.geometric}
\usepackage{pgfplots}
\pgfplotsset{compat=1.18}

\usepackage{hyperref}
\hypersetup{
  colorlinks=true,
  linkcolor=black,
  citecolor=black,
  urlcolor=blue!55!black,
  pdfborder={0 0 0},
  pdftitle={Counterfactual Likelihood Tests for Indirect Influence in Private Reasoning Channels},
  pdfauthor={Alexander Boesgaard Lorup}
}
\usepackage[capitalise]{cleveref}

\usepackage{titlesec}
\titleformat{\section}{\normalfont\large\bfseries}{\thesection}{0.6em}{}
\titleformat{\subsection}{\normalfont\normalsize\bfseries}{\thesubsection}{0.6em}{}
\titlespacing*{\section}{0pt}{1.2\baselineskip}{0.4\baselineskip}
\titlespacing*{\subsection}{0pt}{0.8\baselineskip}{0.25\baselineskip}

\newcommand{\code}[1]{\texttt{#1}}
\newcommand{\Apriv}{\code{A\_private}}
\newcommand{\Apub}{\code{A\_public}}
\newcommand{\Bpriv}{\code{B\_private}}
\newcommand{\Bpub}{\code{B\_public}}

\newcommand{\Cnat}{C_{\mathrm{nat}}}
\newcommand{\Ccf}{C_{\mathrm{cf}}}

\title{\bfseries Counterfactual Likelihood Tests for Indirect Influence\\in Private Reasoning Channels}

\author{Alexander Boesgaard Lorup\\
\emph{Openhagen}}

\date{May 14, 2026}

\begin{document}
\maketitle

\begin{abstract}
\noindent
Reasoning systems increasingly separate intermediate computation into channels: visible chain-of-thought, debate transcripts, verifier notes, private scratchpads, and role-specific deliberation. Once a system has both private and public channels, evaluation must distinguish three cases that look similar in transcripts: independent co-derivation, direct access to private content, and indirect influence through public communication. This paper presents a counterfactual likelihood test for measuring influence between private reasoning channels. The method replaces an upstream private block with a donor block, holds the public token sequence and downstream target fixed, and measures the downstream target's negative-log-likelihood shift.

On a 7B role-channel reasoning model used for validation, simple textual probes fail. Raw 4-gram overlap reports $84\%$ leakage on a masked variant because roles independently derive the same equations and facts. Corrected $n$-gram overlap remains noisy, with only a 10-point gap between unmasked and masked conditions. Canary probes report $0\%$ reproduction in both conditions because instruction-tuned models often ignore planted nonsense even when it is visible. Counterfactual likelihood separates the conditions: an unmasked baseline shows a $54.4\%$ influence rate, while a masked variant shows $13.6\%$.

A second validation step controls a RoPE positional confound by matching donor and original private-block token lengths. The hardened validator shows a near-zero reverse channel (B-to-A $2.0\%$, A-to-B $25.3\%$ on initial pilot validation). A multi-checkpoint validation across three checkpoints from the same training lineage, five random seeds, and 500 sampled traces per checkpoint-seed pair replicates the directional asymmetry at scale: A-to-B rates of $23.6$\textendash$39.3\%$ versus B-to-A rates of $2.3$\textendash$7.0\%$ at the substantive threshold $\Delta_{\mathrm{NLL}} > 0.05$ across 13{,}734 valid directional contrasts. A graph-separation control that blocks private-to-public carrier edges produces bit-identical natural and counterfactual scores across all 13{,}734 control evaluations, identifying the tested public-channel pathway as the complete carrier of the measured counterfactual signal under the implemented role-visibility mask.

The interpretation is not that the attention mask simply fails. Rather, direct reverse access is effectively blocked, while A-to-B influence persists through public-speech hidden states: the public utterance encodes a computational trace of the private reasoning that produced it, and a downstream role allowed to read that public utterance is influenced indirectly. Private-channel evaluation should therefore report direct and indirect influence separately. Counterfactual likelihood probes provide a practical default for this measurement problem.
\end{abstract}

\section{Introduction}

Private reasoning channels are useful only if their information boundaries can be measured. In role-structured reasoning, one component may reason privately before making a public claim; another component may verify that claim without direct access to the first component's scratchpad. This setup raises a concrete question:

\begin{quote}
\itshape
When the verifier's continuation is likely under the proposer-specific private context, is that because the verifier saw private information, because it read the public utterance, or because both roles independently solved the same problem?
\end{quote}

Surface text cannot answer this reliably. Two roles solving the same problem may share units, equations, answer-option strings, and named concepts without any leakage. Conversely, a role that can see a planted random string may decline to repeat it because the string is irrelevant or obviously false. A useful metric must intervene on the upstream channel and score the downstream channel directly.

The validation experiments use a 7B-parameter autoregressive language model with role-separated private and public reasoning channels. The method itself applies to any autoregressive model that exposes or simulates multiple reasoning channels. The relevant channels are:

\begin{itemize}
\setlength{\itemsep}{1pt}
\item \Apriv: proposer private reasoning.
\item \Apub: proposer public utterance.
\item \Bpriv: verifier private reasoning.
\item \Bpub: verifier public utterance.
\end{itemize}

This paper intentionally focuses on the measurement method. It does not rely on the broader training recipe, downstream policy machinery, or benchmark performance of the validation substrate. Accordingly, architectural and training details not required to define the role-visibility graph, intervention protocol, or scoring procedure are intentionally treated as validation-substrate details rather than as contributions of this paper.

\section{Related Work}

Chain-of-thought prompting shows that intermediate reasoning can improve performance in large language models~\cite{wei2022cot}. Self-consistency improves on a single chain by sampling multiple reasoning paths and selecting the most consistent answer~\cite{wang2022selfconsistency}. Tree of Thoughts extends this idea into explicit search over intermediate thought states~\cite{yao2023tot}. Debate-style systems also rely on structured intermediate exchanges, whether as a safety proposal with a judge~\cite{irving2018debate} or as multi-agent language-model deliberation~\cite{du2023multiagent}.

These methods make intermediate reasoning more prominent, but intermediate reasoning is not automatically faithful. Chain-of-thought explanations can rationalize biased answers~\cite{turpin2023unfaithful}, and CoT faithfulness varies under interventions on the reasoning trace~\cite{lanham2023faithfulness}. Chain-of-thought monitorability is therefore promising but fragile~\cite{korbak2025monitor}.

The counterfactual influence test is closest in spirit to causal-intervention work. Causal mediation analysis has been used to study information flow through neural NLP systems~\cite{vig2020mediation}, and causal tracing has been used to identify activations involved in factual predictions~\cite{meng2022rome}. The method here is simpler and more behavioral: it perturbs prefix content and measures downstream target likelihood.

Recent work has also explored parallel multi-stream architectures with role-typed channels~\cite{su2026multistream}, where streams for user, system, model, thinking, and auxiliary audit roles are decoded simultaneously with stream-aware position encoding and a cross-stream causal attention mask. The architectural goals are efficiency, security, and monitorability; the cross-stream information flow they exhibit is amenable to the same counterfactual-likelihood analysis we present. The methodology generalizes to any role-structured system where channels can be independently substituted while preserving downstream targets.

\section{Counterfactual Influence}

Let $X$ be an upstream private block, $P$ the public token sequence that follows it, and $T$ the downstream target continuation. Let $X'$ be a donor private block from another sample. We compare two prefixes:
\begin{align}
\Cnat &= [\mathrm{context},\, X,\, P], \\
\Ccf  &= [\mathrm{context},\, X',\, P].
\end{align}

For a model $p_\theta$, define the mean target negative log-likelihood:
\begin{equation}
\ell(C, T) \;=\; -\frac{1}{|T|} \sum_{i=1}^{|T|} \log p_\theta\!\left(t_i \,\middle|\, C,\, t_{<i}\right).
\label{eq:nll}
\end{equation}

The influence delta is:
\begin{equation}
\Delta_{X \rightarrow T} \;=\; \ell(\Ccf,\, T) \;-\; \ell(\Cnat,\, T).
\label{eq:delta}
\end{equation}

If $\Delta_{X \rightarrow T} > 0$, the natural private block made the downstream target more likely than the donor private block. The \emph{influence rate} is the fraction of valid probes with positive delta. Mean delta is reported in nats/token.

\begin{figure}[t]
\centering
\begin{tikzpicture}[
  font=\small,
  >={Stealth[length=4pt]},
  box/.style={draw, rounded corners=2pt, minimum height=9mm, inner sep=4pt,
              align=center, fill=gray!4},
  priv/.style={box, fill=blue!6, draw=blue!50!black, minimum width=20mm},
  pub/.style={box, fill=orange!10, draw=orange!60!black, minimum width=22mm},
  tgt/.style={box, fill=green!8, draw=green!50!black, minimum width=18mm},
  nll/.style={box, fill=gray!8, minimum width=42mm, minimum height=8mm},
  lab/.style={font=\scriptsize\itshape, gray}
]

\node[box, minimum width=20mm] (prob) at (0, 0) {Problem};
\node[priv] (xnat) at (3.2, 0.9)  {$X$\\\scriptsize natural};
\node[priv] (xcf)  at (3.2, -0.9) {$X'$\\\scriptsize donor, same length};
\node[pub]  (apub) at (7.3, 0) {$P$ \scriptsize(\Apub)};
\node[tgt]  (target) at (10.7, 0) {$T$ \scriptsize(target)};

\draw[->] (prob.east) -- (xnat.west);
\draw[->] (prob.east) -- (xcf.west);
\draw[->] (xnat.east) -- (apub.west);
\draw[->, dashed] (xcf.east) -- (apub.west);
\draw[->] (apub.east) -- (target.west);

\node[lab] at (3.2, 1.95) {upstream private block is replaced};

\draw[gray!40, dashed] (-0.6, -2.0) -- (12.2, -2.0);

\node[nll] (lnat) at (3.0, -2.9) {$\ell_{\mathrm{nat}} = \ell(\Cnat, T)$};
\node[nll] (lcf)  at (8.8, -2.9) {$\ell_{\mathrm{cf}}  = \ell(\Ccf,  T)$};
\draw[->] (lnat.east) -- (lcf.west);

\node[font=\small] at (5.9, -4.0)
  {$\Delta_{X\rightarrow T} = \ell_{\mathrm{cf}} - \ell_{\mathrm{nat}};
    \quad \Delta > 0$ indicates upstream influence};

\end{tikzpicture}
\caption{Counterfactual influence test design. Public tokens and the target continuation are held fixed; only the upstream private block is replaced.}
\label{fig:design}
\end{figure}
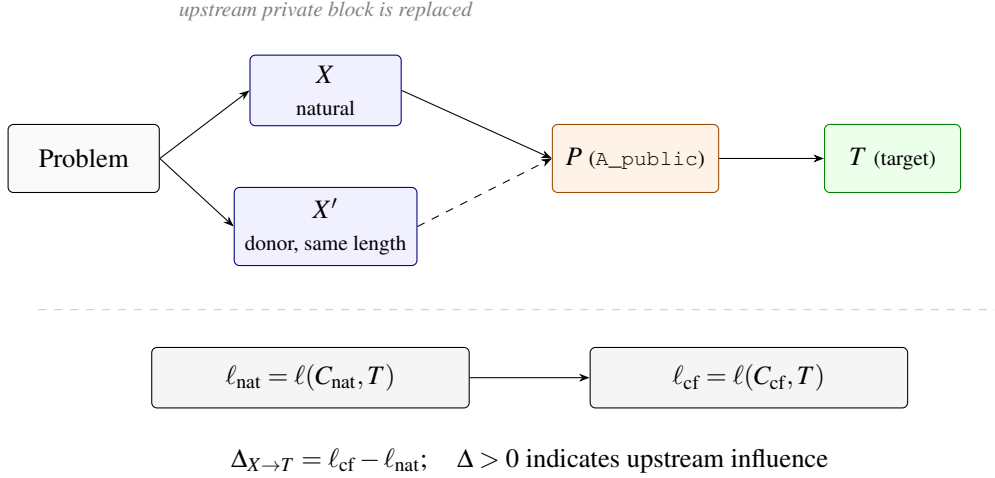

The target continuation is fixed. The test does not ask whether the downstream role would regenerate the same text from scratch. It asks whether the observed downstream target is better explained by the natural upstream private block than by a counterfactual private block.

\section{Probe Evolution}

Three simpler probes were evaluated before adopting counterfactual likelihood as the primary metric.

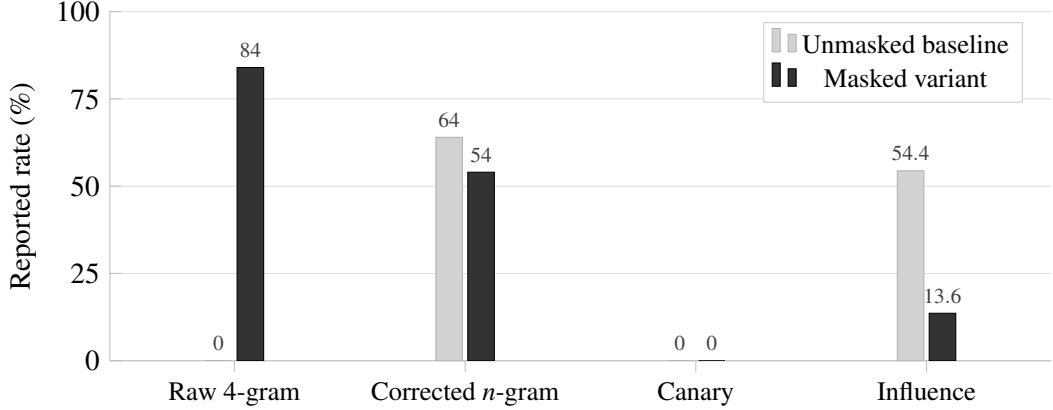
\begin{figure}[t]
\centering
\begin{tikzpicture}
\begin{axis}[
  width=0.85\linewidth, height=6.2cm,
  ybar=2pt,
  bar width=10pt,
  enlarge x limits=0.18,
  ymin=0, ymax=100,
  ylabel={Reported rate (\%)},
  symbolic x coords={Raw 4-gram, Corrected $n$-gram, Canary, Influence},
  xtick=data,
  xticklabel style={align=center, font=\small},
  ytick={0,25,50,75,100},
  ymajorgrids=true, grid style={gray!25},
  axis line style={gray!50},
  tick style={gray!50},
  legend style={
    at={(0.97,0.97)}, anchor=north east,
    draw=gray!40, font=\small, fill=white,
    /tikz/every even column/.append style={column sep=8pt}
  },
  nodes near coords,
  nodes near coords style={font=\scriptsize, color=black!75},
  axis x line*=bottom, axis y line*=left,
]
\addplot+[fill=gray!35, draw=gray!60] coordinates {
  (Raw 4-gram, 0)
  (Corrected $n$-gram, 64)
  (Canary, 0)
  (Influence, 54.4)
};
\addplot+[fill=black!80, draw=black] coordinates {
  (Raw 4-gram, 84)
  (Corrected $n$-gram, 54)
  (Canary, 0)
  (Influence, 13.6)
};
\legend{Unmasked baseline, Masked variant}
\end{axis}
\end{tikzpicture}
\caption{Probe families differ sharply in diagnostic value. Raw 4-gram overlap was measured only on the masked variant.}
\label{fig:probes}
\end{figure}

\subsection{Raw $n$-gram overlap}

The first probe intersected 4-grams between \Apriv{} and B's downstream private/public target. It reported $84\%$ leakage on a masked model. Manual inspection showed that the overlap was largely legitimate co-derivation: two roles solving the same problem independently used the same formulas, terms, answer-option text, and intermediate facts. Raw overlap therefore overstates leakage.

\subsection{Corrected $n$-gram overlap}

The second probe subtracted $n$-grams already present in the prompt and prior public utterances. This moved in the right direction but remained too noisy. The unmasked baseline scored $64\%$, while the masked variant scored $54\%$. A 10-point gap is useful as a diagnostic but too weak for a primary channel-isolation test.

\subsection{Canary reproduction}

The third probe inserted random strings or absurd claims into a private block and checked whether the downstream role reproduced them. It reported $0\%$ reproduction in both unmasked and masked settings. This does not prove absence of access. It shows that instruction-tuned reasoning models often refuse to repeat irrelevant nonsense, even when the content is visible.

\subsection{Counterfactual likelihood}

The fourth probe scored the same downstream target under the natural and counterfactual prefixes. This separated the conditions cleanly:

\begin{table}[h]
\centering
\small
\begin{tabular}{lccl}
\toprule
Probe & Unmasked baseline & Masked variant & Interpretation \\
\midrule
Corrected $n$-gram overlap & $64.0\%$ & $54.0\%$ & Directional but noisy \\
Canary reproduction        & $0.0\%$  & $0.0\%$  & No discrimination \\
Counterfactual influence   & $\mathbf{54.4\%}$ & $\mathbf{13.6\%}$ & Strong discrimination \\
\bottomrule
\end{tabular}
\caption{Probe comparison on the unmasked baseline and masked variant.}
\label{tab:probes}
\end{table}

The unmasked baseline mean delta was $0.068$ nats/token. The masked variant mean delta was $0.022$ nats/token. Counterfactual likelihood was the first probe that both separated the mask conditions and avoided treating shared problem-derived text as leakage.

\section{RoPE-Hardened Validation}

The first counterfactual implementation exposed a positional confound. If the donor private block has a different token count from the original block, all subsequent tokens shift position. In a RoPE-based transformer, this can change downstream likelihood even when the target role cannot attend to the replaced private tokens. The likelihood shift can therefore measure position, not content.

The symptom was an A-to-B influence rate of $22.2\%$ despite blocked direct access from B to A's private tokens. The top-delta examples had large donor/original length mismatches, often with unrelated donor domains. B-to-A was only $5.1\%$, already suggesting that reverse influence was near zero.

The hardened validator controls this by using exact token-length-matched donor private blocks. After length matching, the validation produced:

\begin{table}[h]
\centering
\small
\begin{tabular}{lc}
\toprule
Metric & Value \\
\midrule
Valid traces                & $99/100$ \\
Overall influence rate      & $13.6\%$ \\
A-to-B influence rate       & $25.3\%$ \\
A-to-B mean delta           & $0.038$ nats/token \\
B-to-A influence rate       & $2.0\%$ \\
B-to-A mean delta           & $0.011$ nats/token \\
Diagnostic $n$-gram overlap & $57.0\%$ \\
\bottomrule
\end{tabular}
\caption{Hardened (length-matched) validation on the masked variant.}
\label{tab:hardened}
\end{table}

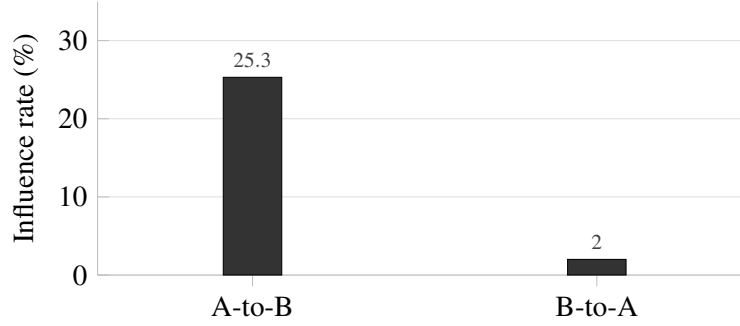
\begin{figure}[t]
\centering
\begin{tikzpicture}
\begin{axis}[
  width=0.62\linewidth, height=5.2cm,
  ybar=2pt,
  bar width=22pt,
  enlarge x limits=0.45,
  ymin=0, ymax=35,
  ylabel={Influence rate (\%)},
  symbolic x coords={A-to-B, B-to-A},
  xtick=data,
  ymajorgrids=true, grid style={gray!25},
  axis line style={gray!50},
  tick style={gray!50},
  nodes near coords,
  nodes near coords style={font=\scriptsize, color=black!75},
  axis x line*=bottom, axis y line*=left,
]
\addplot+[fill=black!80, draw=black] coordinates {
  (A-to-B, 25.3) (B-to-A, 2.0)
};
\end{axis}
\end{tikzpicture}
\caption{Directional influence after length matching on the masked variant. The reverse channel is near zero; A-to-B influence persists through public-speech hidden states.}
\label{fig:directional}
\end{figure}

The hardened result supports a directional interpretation. B-to-A influence is near zero, consistent with a blocked reverse private channel. A-to-B influence remains non-zero through a separate mechanism: public-speech hidden states retain a computational trace of the private content that preceded them, even when the public token IDs themselves are unchanged. Downstream roles allowed to attend to public communication can therefore receive private-derived information indirectly, without violating any private-token attention mask.

This distinction is essential. A direct private-token mask and an indirect public-hidden-state channel are not the same mechanism. The first is a channel-isolation failure. The second may be architecturally intended, depending on whether public speech is allowed to carry privately reasoned information.

\section{Strengthened Multi-Checkpoint Validation}

The hardened pilot used roughly $100$ valid traces from a single checkpoint, which is sufficient to expose the RoPE confound and identify the directional pattern, but insufficient to establish that the asymmetry holds at scale or across training variants. To test robustness, we ran an extended validation across three checkpoints from the same training lineage, five random seeds per checkpoint, and 500 sampled traces per checkpoint-seed pair.

\subsection{Scale}

For each checkpoint and seed, we sampled 500 traces. Each valid trace could contribute directional A-to-B and B-to-A contrasts, subject to donor availability and length matching. The full validation comprised $7{,}500$ sampled traces ($3$ checkpoints $\times$ $5$ seeds $\times$ $500$ samples), of which $7{,}298$ were valid after trace-level filtering (overall validity $97.3\%$). After donor-availability and length-matching constraints, valid traces yielded $13{,}734$ valid directional contrasts (sum of A-to-B and B-to-A across checkpoints) and a matched number of graph-cut control evaluations. Counting both natural and counterfactual prefix/target evaluations, the validation involved approximately $54{,}900$ teacher-forced likelihood computations.

\subsection{Multi-Checkpoint Directional Asymmetry}

Across all three checkpoints, length-matched counterfactual replacement of \Apriv{} consistently increased the negative log-likelihood of the immediate B-attributed continuation window relative to the natural prefix. The reverse direction\textemdash\Bpriv{} replacement, measured against the next A-attributed continuation under the role-visibility graph\textemdash remained substantially smaller in every checkpoint.

\paragraph{A clarifying note on directionality.} The channel ordering introduced in Section 1 describes the within-trace temporal flow of a single proposer-verifier exchange: \Apriv{} precedes \Apub, which precedes \Bpriv, which precedes \Bpub. The directional contrasts here are not backwards-in-time interventions on a flattened causal sequence; they are reverse role-direction contrasts under the role-visibility graph.

For A-to-B, the prefix is the prompt followed by the first \Apriv{} and \Apub, and the target is the first \Bpriv{} and \Bpub. This contrast is both temporally and graph-wise upstream-to-downstream.

For B-to-A, the substituted block is \Bpriv, and the target is the next A-attributed continuation under the role-visibility graph. When a sampled trace contains a natural second A-attributed continuation following the first exchange (and any intermediate public outputs the graph permits), that continuation is the target. When the sampled trace ends after one exchange, the validator appends a fixed continuation bridge to the natural prefix, generates a next-A target under that natural prefix, and then teacher-forces the same target under both the natural and length-matched \Bpriv-replaced prefixes. Across the three checkpoints, the synthetic-bridge fallback was used in $82$\textendash$95\%$ of B-to-A contrasts (with proportions varying by checkpoint due to differences in natural multi-turn frequency); the remaining B-to-A contrasts used natural second-A continuations. The reported B-to-A rates pool both cases.

Thus the B-to-A contrast measures whatever reverse role-direction influence the role-visibility graph permits, not literal backwards-in-time influence on the already-generated first \Apub. Under a strictly blocked reverse-private channel and no graph paths from \Bpriv{} to any subsequent A-attributed continuation, the B-to-A rate should approach zero; observed residuals therefore indicate either residual mask leakage or permitted public-channel routing paths. The graph-cut control in Section~\ref{ssec:graphcut} disambiguates these two cases.

\Cref{tab:multicheckpoint} reports rates at the substantive threshold $\Delta_{\mathrm{NLL}} > 0.05$, chosen to filter for clearly positive influence above the noise floor implied by mean delta magnitudes.

\begin{table}[h]
\centering
\small
\begin{tabular}{lcccc}
\toprule
Checkpoint & A-to-B rate (95\% CI) & A-to-B mean $\Delta_{\mathrm{NLL}}$ & B-to-A rate (95\% CI) & B-to-A mean $\Delta_{\mathrm{NLL}}$ \\
\midrule
Checkpoint 1 & $23.6\%$ \scriptsize{[$21.9$, $25.4$]} & $0.034$ & $2.3\%$ \scriptsize{[$1.8$, $3.0$]} & $0.011$ \\
Checkpoint 2 & $39.0\%$ \scriptsize{[$37.0$, $41.0$]} & $0.048$ & $7.0\%$ \scriptsize{[$6.0$, $8.1$]} & $0.017$ \\
Checkpoint 3 & $39.3\%$ \scriptsize{[$37.3$, $41.3$]} & $0.048$ & $6.9\%$ \scriptsize{[$5.9$, $8.0$]} & $0.018$ \\
\bottomrule
\end{tabular}
\caption{Multi-checkpoint validation at substantive threshold $\Delta_{\mathrm{NLL}} > 0.05$. Rates are reported with Wilson 95\% confidence intervals computed from the underlying counts (Checkpoint 1: $546/2311$ A-to-B, $54/2348$ B-to-A; Checkpoint 2: $886/2273$ A-to-B, $160/2279$ B-to-A; Checkpoint 3: $881/2244$ A-to-B, $157/2279$ B-to-A). All three checkpoints are masked variants from the same training lineage, each evaluated under five random seeds with 500 sampled traces per checkpoint-seed pair.}
\label{tab:multicheckpoint}
\end{table}

The directional asymmetry holds in every checkpoint. The A-to-B and B-to-A 95\% confidence intervals are non-overlapping by a wide margin in all three checkpoints, indicating that the asymmetry is not attributable to sample-size noise. A-to-B mean delta exceeds B-to-A mean delta by $0.022$ to $0.032$ nats/token across the three checkpoints. A-to-B influence rate is $5.6$ to $10.3$ times the corresponding B-to-A rate. Absolute magnitudes shift across checkpoints, but the asymmetric pattern is consistent.

\paragraph{Clustering check.} Because the directional contrasts are clustered by checkpoint, seed, trace, task, and donor pool, the Wilson 95\% confidence intervals above treat contrasts as independent Bernoulli samples and may understate uncertainty. As a descriptive clustering check, we computed influence rates and mean $\Delta_{\mathrm{NLL}}$ independently for each of the five seeds per checkpoint and compared A-to-B against B-to-A within every seed-checkpoint cell. A-to-B exceeded B-to-A on both metrics in all $15$ seed-checkpoint cells. Treating cells as exchangeable under a no-preference null gives a one-sided sign-test probability of $2^{-15} \approx 3.05 \times 10^{-5}$ for the joint cell-level comparison ($2^{-5} = 0.03125$ per checkpoint). This is a descriptive sanity check, not a formal hypothesis test, since seed-checkpoint cells are not strictly independent. Per-checkpoint seed-level means and standard deviations are reported in Appendix~A.

\begin{figure}[t]
\centering
\begin{tikzpicture}
\begin{axis}[
  width=0.78\linewidth, height=5.6cm,
  ybar=2pt,
  bar width=14pt,
  enlarge x limits=0.25,
  ymin=0, ymax=45,
  ylabel={Influence rate at $\Delta_{\mathrm{NLL}} > 0.05$ (\%)},
  xlabel={},
  symbolic x coords={Checkpoint 1, Checkpoint 2, Checkpoint 3},
  xtick=data,
  ymajorgrids=true, grid style={gray!25},
  axis line style={gray!50},
  tick style={gray!50},
  legend style={
    at={(0.97,0.97)}, anchor=north east,
    draw=gray!40, font=\small, fill=white
  },
  nodes near coords,
  nodes near coords style={font=\scriptsize, color=black!75},
  axis x line*=bottom, axis y line*=left,
]
\addplot+[fill=black!80, draw=black] coordinates {
  (Checkpoint 1, 23.6) (Checkpoint 2, 39.0) (Checkpoint 3, 39.3)
};
\addplot+[fill=gray!35, draw=gray!60] coordinates {
  (Checkpoint 1, 2.3)  (Checkpoint 2, 7.0)  (Checkpoint 3, 6.9)
};
\legend{A-to-B, B-to-A}
\end{axis}
\end{tikzpicture}
\caption{Directional influence rates across three checkpoints from the same training lineage. A-to-B influence exceeds B-to-A by a factor of $5.6$ to $10.3$ in every checkpoint; absolute magnitudes vary, the asymmetric pattern is consistent.}
\label{fig:multicheckpoint}
\end{figure}
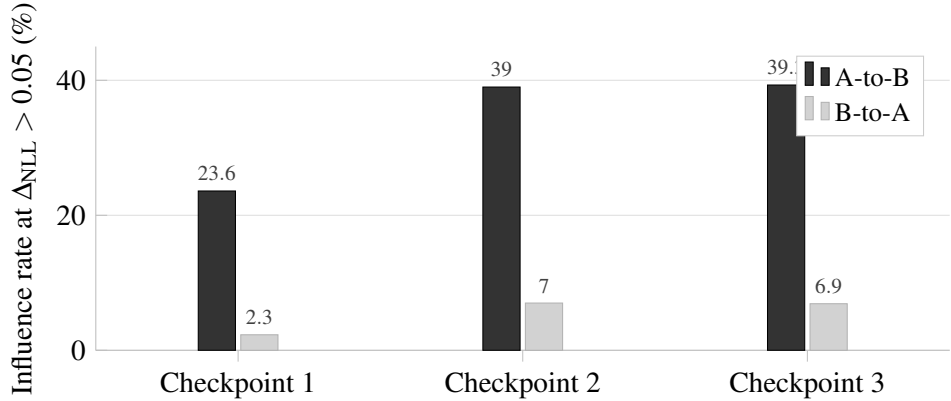

\subsection{Graph-Cut Control for Private-to-Public Carrier Edges}
\label{ssec:graphcut}

The hardened result above is consistent with two interpretations: the indirect influence flows through public-speech hidden states, or it flows through some other channel we have not isolated. To distinguish these, we ran a graph-cut control: we re-scored the same prefix and target under a modified attention pattern that blocks specific private-to-public carrier edges while leaving all other public-channel attention intact.

\paragraph{Purpose.} This control is not a deployment-mode text-only baseline; it is a graph-separation intervention used to test whether the specified carrier edges are sufficient to transmit the counterfactual signal. The point of the control is to validate the graph path that carries indirect influence, not to simulate ordinary model use. An exact-zero result is partly by construction\textemdash if the blocked edges are the carrier channel, computation downstream of the cut is identical between natural and counterfactual prefixes\textemdash and identifying this constructive property is the methodological contribution.

The control is designed such that, after blocking the specified edges, the substituted private span has no path to the scored target tokens under the role-visibility mask. If the specified edges \emph{are} the carrier channel, then by construction the natural and counterfactual prefixes produce identical hidden states at every position in the target window, and the resulting target NLL is bit-identical. If unblocked edges still carry the channel, the natural and counterfactual scores will diverge.

\begin{table}[h]
\centering
\small
\begin{tabular}{lccc}
\toprule
Checkpoint & Control evaluations & $\Delta_{\mathrm{NLL}} = 0$ rate & Mean $\Delta_{\mathrm{NLL}}$ \\
\midrule
Checkpoint 1 & $4{,}659$  & $100.0\%$ & $0.000$ \\
Checkpoint 2 & $4{,}552$  & $100.0\%$ & $0.000$ \\
Checkpoint 3 & $4{,}523$  & $100.0\%$ & $0.000$ \\
\midrule
Total        & $13{,}734$ & $100.0\%$ & $0.000$ \\
\bottomrule
\end{tabular}
\caption{Graph-cut control across the three checkpoints. Across $13{,}734$ length-matched probes pooling both directional pairs (A-to-B and B-to-A), natural and counterfactual scores were bit-identical in every case after blocking the private-to-public carrier edges.}
\label{tab:graphsep}
\end{table}

The exact-zero result is the structural signature of complete graph separation. It is not produced by measurement-noise reduction: the underlying NLL values in the control range across roughly $0.05$\textendash$0.90$ across probes (mean approximately $0.4$, standard deviation approximately $0.14$), and natural-scores under the control differ from natural-scores in the primary condition in over $99\%$ of probes. The exact zero arises because the modified attention pattern places the substituted private span causally outside any path to the scored target tokens; computation downstream of the cut is therefore identical between natural and counterfactual prefixes.

This is a stronger result than ``the indirect channel is consistent with public hidden states.'' Under the implemented role-visibility mask, the tested private-to-public carrier edges form a complete path set for the measured counterfactual signal: blocking these edges eliminates the signal by construction. The methodology can therefore identify not just \emph{that} an indirect channel exists, but \emph{which specific edges} carry it.

\subsection{Interpretation Across Checkpoints}

The combination of multi-checkpoint replication and the graph-separation control supports a specific mechanistic picture:

\emph{First,} the direct private-token attention mask is approximately effective. B-to-A influence is substantially smaller than A-to-B in every checkpoint, and the graph-cut control eliminates the B-to-A residual as well, indicating that the small B-to-A residual flows through public-channel paths rather than through direct private-token access.

\emph{Second,} the public-speech hidden-state channel is the dominant pathway for A-to-B influence. Blocking the private-to-public carrier edges reduces A-to-B influence to zero across all observations and all three checkpoints.

\emph{Third,} the methodology has discrimination power across training variants. Absolute influence rates vary across checkpoints by a factor of approximately $1.7$ for A-to-B and approximately $3$ for B-to-A while preserving the directional pattern, indicating that the metric is sensitive to training-induced changes in channel magnitude without losing the qualitative asymmetry.

The graph-separation result is architecture-specific. It identifies the carrier edges for the role-visibility mask used here; systems with different mask topologies will have different candidate edge sets and may require correspondingly different controls. The principle generalizes, but the specific result does not.

\section{Discussion}

The counterfactual influence test makes three distinctions that textual probes miss.

\emph{First,} independent co-derivation is not leakage. Reasoning tasks naturally create shared surface text. A role that derives the same equation from the same prompt has not necessarily read another role's private scratchpad.

\emph{Second,} absence of canary reproduction is not absence of access. Instruction-tuned models can see nonsense and choose not to repeat it.

\emph{Third,} public communication is a channel. If public hidden states carry private-derived information, a downstream role can be influenced without direct private-token access. This may be acceptable, but it must be measured and named.

The method is not a full mechanistic proof. It does not identify which layer, head, or activation subspace carries the signal. It is a behavioral intervention over prefix content. Activation patching, causal mediation, and path-specific analyses can refine the mechanism after the behavioral channel has been detected.

\section{Recommendations}

\textbf{Report directionality.} A single influence score hides role-order effects. A-to-B and B-to-A should be reported separately.

\textbf{Separate direct and indirect channels.} Private-token attention, public token IDs, public hidden states, and final answer extraction are distinct surfaces.

\textbf{Length-match donors.} In RoPE models, donor length mismatch can create false influence through positional shifts.

\textbf{Treat $n$-grams and canaries as diagnostics.} They are useful for debugging but insufficient as primary gates.

\textbf{Track influence after training changes.} Fine-tuning can amplify or reduce indirect channels without changing the formal attention mask.

\textbf{Define the allowed boundary.} Some applications want public speech to transmit private reasoning; others want public speech to be re-encoded through a bottleneck. The metric should be interpreted relative to that policy.

\section{Limitations}

The reported results come from one role-channel reasoning architecture family, evaluated across three checkpoints from the same training lineage. The absolute influence rates may not transfer to other model families, role layouts, or masking implementations. The graph-separation control identifies the carrier edges for the role-visibility mask used here; systems with different mask topologies may have additional channels the present control would not block. The methodology generalizes\textemdash for any candidate channel, a corresponding graph-separation control can be constructed\textemdash but the specific exact-zero result is architecture-dependent.

The reported influence rates apply to the specific checkpoints evaluated. Training interventions that shape downstream-role behavior (e.g., toward more independent verification or different reliance patterns on upstream public statements) may attenuate or amplify indirect channels. The methodology measures influence at evaluation time; it does not characterize how training choices affect channel magnitudes.

Because most B-to-A contrasts use a fixed continuation bridge rather than naturally occurring second-A continuations ($82$\textendash$95\%$ synthetic-bridge fallback across checkpoints, as documented in Section~6.2), B-to-A should be interpreted as a reverse role-direction control under the visibility graph, not as a fully symmetric conversational-turn analogue of A-to-B. The directional comparison is well-defined under the role-visibility graph but is not a perfectly matched within-trace counterpart of A-to-B.

The decision thresholds used are simple: $\Delta_{\mathrm{NLL}} > 0$ for inclusive influence and $\Delta_{\mathrm{NLL}} > 0.05$ for substantive influence. Future work could add clustered bootstrap intervals, finer threshold sweeps, donor-domain stratification, target-length controls, and semantic similarity controls.

The test measures likelihood influence, not harm. A high A-to-B rate may be acceptable when B is supposed to read A's public argument. A low B-to-A rate may be reassuring for one role order but not for another. Influence metrics should be interpreted alongside the system's intended information policy.

\section{Conclusion}

Private reasoning channels require causal-style evaluation. Text overlap overstates influence because independent roles share problem-derived content. Canary reproduction understates influence because instruction-tuned models may ignore planted nonsense. Counterfactual likelihood is a better first-line probe: replace the upstream private block, hold public tokens and downstream target fixed, and measure whether the target becomes less likely.

On the role-channel architecture used here, this probe separated unmasked from masked behavior, exposed a RoPE positional confound, and identified the real residual channel: A-to-B influence through public-speech hidden states, with near-zero reverse influence. A multi-checkpoint validation across three checkpoints, five seeds, and $13{,}734$ valid directional contrasts replicates the directional asymmetry at scale. A graph-separation control that blocks private-to-public carrier edges produces bit-identical natural and counterfactual scores in $100\%$ of $13{,}734$ control evaluations, identifying the tested public-channel pathway as the complete carrier of the measured counterfactual signal under the implemented role-visibility mask.

The practical lesson is to specify and measure the boundary precisely. Private tokens, public tokens, public hidden states, and final answers are different channels; counterfactual influence tests, combined with graph-separation controls on specific carrier edges, make those distinctions observable.

\appendix
\section*{Appendix A: Seed-Level Statistics}
\addcontentsline{toc}{section}{Appendix A: Seed-Level Statistics}

For each checkpoint, the table below reports the seed-level mean and standard deviation of the A-to-B and B-to-A influence rates (at substantive threshold $\Delta_{\mathrm{NLL}} > 0.05$) and the seed-level mean and standard deviation of the per-condition mean $\Delta_{\mathrm{NLL}}$. Each per-checkpoint statistic is computed across five random seeds ($n=5$). The right-most columns report the count of seed-checkpoint cells in which A-to-B exceeds B-to-A.

\begin{table}[H]
\centering
\footnotesize
\renewcommand{\arraystretch}{1.15}
\begin{tabular}{lcccccc}
\toprule
 & A-to-B rate & B-to-A rate & A-to-B $\Delta_{\mathrm{NLL}}$ & B-to-A $\Delta_{\mathrm{NLL}}$ & A$>$B & A$>$B \\
Checkpoint & mean $\pm$ SD (\%) & mean $\pm$ SD (\%) & mean $\pm$ SD & mean $\pm$ SD & by rate & by $\Delta$ \\
\midrule
Checkpoint 1 & $23.61 \pm 3.60$ & $2.30 \pm 0.69$ & $0.0338 \pm 0.0017$ & $0.0113 \pm 0.0005$ & $5/5$ & $5/5$ \\
Checkpoint 2 & $38.98 \pm 1.89$ & $7.02 \pm 1.13$ & $0.0481 \pm 0.0010$ & $0.0165 \pm 0.0011$ & $5/5$ & $5/5$ \\
Checkpoint 3 & $39.25 \pm 1.21$ & $6.89 \pm 0.64$ & $0.0478 \pm 0.0020$ & $0.0175 \pm 0.0011$ & $5/5$ & $5/5$ \\
\midrule
All cells    & \multicolumn{4}{c}{(combined across checkpoints)} & $15/15$ & $15/15$ \\
\bottomrule
\end{tabular}
\caption{Per-checkpoint seed-level statistics. Rates are influence rates at $\Delta_{\mathrm{NLL}} > 0.05$, expressed as percentages. Each cell summary is computed across the five random seeds for that checkpoint. The two right-most columns count seed-checkpoint cells in which the A-to-B value exceeds the B-to-A value (by rate or by mean $\Delta_{\mathrm{NLL}}$, respectively).}
\label{tab:seedlevel}
\end{table}

A-to-B exceeds B-to-A in every one of the $15$ seed-checkpoint cells, on both rate and mean-$\Delta_{\mathrm{NLL}}$ metrics. Under a null of no directional preference, the probability of all $15$ cells favoring the same direction is $p = 2^{-15} \approx 3.05 \times 10^{-5}$. Within-checkpoint, the per-seed standard deviation of the A-to-B rate ranges from $1.21$ to $3.60$ percentage points; the per-seed standard deviation of A-to-B mean $\Delta_{\mathrm{NLL}}$ ranges from $0.0010$ to $0.0020$ nats/token. These within-checkpoint variances are an order of magnitude smaller than the between-direction gaps reported in \Cref{tab:multicheckpoint}, supporting the interpretation that the directional asymmetry is robust to seed clustering.

\section*{Appendix B: Citation Verification}
\addcontentsline{toc}{section}{Appendix B: Citation Verification}

The following papers were checked against primary arXiv pages during preparation. The table states only the claims this paper relies on.

\begin{table}[H]
\centering
\footnotesize
\renewcommand{\arraystretch}{1.15}
\begin{tabularx}{\linewidth}{@{}p{0.30\linewidth} X l@{}}
\toprule
Citation & Claim used here & arXiv \\
\midrule
Wei et al., 2022, ``Chain-of-Thought Prompting'' &
CoT uses intermediate reasoning steps and improves arithmetic, commonsense, and symbolic reasoning in sufficiently large models. &
\href{https://arxiv.org/abs/2201.11903}{2201.11903} \\

Wang et al., 2022, ``Self-Consistency'' &
Self-consistency samples diverse reasoning paths and selects/marginalizes toward the most consistent answer. &
\href{https://arxiv.org/abs/2203.11171}{2203.11171} \\

Yao et al., 2023, ``Tree of Thoughts'' &
Tree of Thoughts explores coherent intermediate thought units with deliberate search. &
\href{https://arxiv.org/abs/2305.10601}{2305.10601} \\

Irving et al., 2018, ``AI Safety via Debate'' &
Debate proposes agents arguing before a judge as a mechanism for eliciting truthful/useful answers. &
\href{https://arxiv.org/abs/1805.00899}{1805.00899} \\

Du et al., 2023, ``Multiagent Debate'' &
Multiple language-model instances debate over rounds, with reported improvements in reasoning and factuality. &
\href{https://arxiv.org/abs/2305.14325}{2305.14325} \\

Turpin et al., 2023, ``Don't Always Say What They Think'' &
CoT explanations can be unfaithful and rationalize biased answers. &
\href{https://arxiv.org/abs/2305.04388}{2305.04388} \\

Lanham et al., 2023, ``Measuring Faithfulness'' &
Faithfulness can be studied by intervening on chain-of-thought content and observing answer changes. &
\href{https://arxiv.org/abs/2307.13702}{2307.13702} \\

Vig et al., 2020, ``Causal Mediation Analysis'' &
Causal mediation can analyze information flow through neural NLP components. &
\href{https://arxiv.org/abs/2004.12265}{2004.12265} \\

Meng et al., 2022, ``Locating and Editing'' &
Causal tracing and related interventions can localize activations involved in factual predictions. &
\href{https://arxiv.org/abs/2202.05262}{2202.05262} \\

Korbak et al., 2025, ``CoT Monitorability'' &
CoT monitoring is a safety opportunity but fragile and incomplete. &
\href{https://arxiv.org/abs/2507.11473}{2507.11473} \\

Su et al., 2026, ``Multi-Stream LLMs'' &
Parallel multi-stream architectures decode role-typed channels (user, system, model, thinking, and auxiliary audit) simultaneously with stream-aware position encoding and a cross-stream causal attention mask, targeting efficiency, security, and monitorability. &
\href{https://arxiv.org/abs/2605.12460}{2605.12460} \\
\bottomrule
\end{tabularx}
\end{table}

\FloatBarrier

\end{document}